\documentclass[conference]{IEEEtran}
\IEEEoverridecommandlockouts
\usepackage{cite}
\usepackage{amsmath,amssymb,amsfonts}
\usepackage{algorithmic}
\usepackage{graphicx}
\usepackage{textcomp}
\usepackage{xcolor}
\usepackage{subcaption}
\usepackage{booktabs}
\usepackage{enumitem}
\def\BibTeX{{\rm B\kern-.05em{\sc i\kern-.025em b}\kern-.08em
    T\kern-.1667em\lower.7ex\hbox{E}\kern-.125emX}}
\usepackage{graphicx,xcolor}

\begin{document}


\title{DeepOHeat: Operator Learning-based Ultra-fast Thermal Simulation in 3D-IC Design\\
}


\author{\IEEEauthorblockN{Ziyue Liu$^1$, Yixing Li$^2$, Jing Hu$^3$, Xinling Yu$^1$, Shinyu Shiau$^3$, Xin Ai$^3$, Zhiyu Zeng$^2$ and Zheng Zhang$^1$}
\IEEEauthorblockA{
$^1$University of California at Santa Barbara, Santa Barbara, CA. Email: \{ziyueliu, xyu644, zzhang01\}@ucsb.edu\\
$^2$ Cadence Design Systems, Austin, TX. Email: \{yixingli, zzeng\}@cadence.com \\
$^3$ Cadence Design Systems, San Jose, CA. Email: \{jinghu, shinyu, nathanai\}@cadence.com }
}

\maketitle

\begin{abstract}
Thermal issue is a major concern in 3D integrated circuit (IC) design. Thermal optimization of 3D IC often requires massive expensive PDE simulations. Neural network-based thermal prediction models can perform real-time prediction for many unseen new designs. However, existing works either solve 2D temperature fields only or do not generalize well to new designs with unseen design configurations (e.g., heat sources and boundary conditions). In this paper, \textit{for the first time}, we propose DeepOHeat, a physics-aware operator learning framework to predict the temperature field of a family of heat equations with multiple parametric or non-parametric design configurations. This framework learns a functional map from the function space of multiple key PDE configurations (e.g., boundary conditions, power maps, heat transfer coefficients) to the function space of the corresponding solution (i.e., temperature fields), enabling fast thermal analysis and optimization by changing key design configurations (rather than just some parameters). We test DeepOHeat on some industrial design cases and compare it against Celsius 3D from Cadence Design Systems. Our results show that, for the unseen testing cases, a well-trained DeepOHeat can produce accurate results with $1000\times$ to $300000\times$ speedup.
\end{abstract}


\section{Introduction and Related Work}
The increasing transistor density on a silicon chip has led to high power and heat density. The excessive heat can affect the normal performance, reliability, and lifespan of semiconductor chips. Due to the multiple stacked active silicon layers, 3D IC design suffers from much higher power density~\cite{optimal, survey,survey2}. Meanwhile, the increased complexity of 3D chips introduces extra design configurations and system parameters and hence  prolongs the design cycle. Consequently, chip thermal optimization, which provides the optimal thermal-aware floorplan at an early stage, has become an important step in the 3D IC design flow.  Detailed and fast thermal simulators are needed in various thermal-aware design optimization tools.


Discretization-based PDE solvers, such as finite-element and finite-difference methods, have been widely used for 3D chip thermal analysis. The finite-element method (FEM), though computationally expensive, provides the best accuracy and flexibility \cite{survey2}, and is mostly used in commercial solvers such as Celsius, ANSYS, and COMSOL. The finite-difference methods (FDM) are simpler to implement and are widely used in open-source solvers \cite{3dice, efficient-full-chip, ic-thermal}. These thermal simulators provide accurate temperature estimations but cost extensive computational resources. Once a new design is generated, designers need to re-run many simulations to optimize the design case, which can be unaffordable for complicated tasks.  Some surrogate models have been developed to reduce the cost of thermal prediction. For instance, model-order reduction~\cite{wang2004spice,xie2013system} can accelerate each time-domain simulation via reducing the number of state variables in a dynamic system. Data-driven regression methods~\cite{samal2016adaptive,samal2014fast} can model the dependence on certain design parameters in a specified range, but the training step often needs massive high-resolution PDE simulation data. Neither technique can capture the dependence of the temperature field on key PDE configurations (e.g., boundary conditions, non-parametric heat source configurations). 

Neural network-based methods can  perform real-time predictions for unseen data. Several data-driven \cite{dnn-thermal, ml-based-3d, ml-thermal}  and physics-informed neural network-based (PINN) methods \cite{unsupervised-auto-encoder, simnet} have been proposed. However, these existing works either fail to solve 3D full-chip temperature fields, lack generalization to different PDE configurations, or need to be combined with traditional solvers or additional computations. For example, the data-driven method in \cite{dnn-thermal} needs to be combined with a coarse thermal profile obtained by a traditional FEM-based  method. The ML-based transient thermal solver in \cite{ml-based-3d} needs to be combined with convolution operations. The autoencoder-decoder-based methods in \cite{unsupervised-auto-encoder,ml-thermal} are not applicable to 3D volumetric power maps. The PINN-based approach in \cite{simnet} only takes input from geometric parameters rather than general configurations, such as boundary conditions and power maps.

{\bf Paper Contributions.} We propose the DeepOHeat framework, which leverages recent advances in operator learning, as an end-to-end thermal solver for ultra-fast 3D chip thermal prediction under various (both parametric and non-parametric) PDE configurations. Our contributions are as follows:
\begin{itemize}[leftmargin=*]
    \item \textit{For the first time}, an end-to-end operator learning-based 3D IC thermal simulator is proposed to solve a family of heat equations under various PDE configurations.
    \item We propose a modular approach that encodes the PDE configurations of 3D IC designs, including arbitrarily stacked cuboidal geometry, individually defined boundary conditions, 2D/3D power maps, and full-chip flexible material conductivity distribution.
    \item The proposed DeepOHeat achieves $1000\times$ to $300000\times$ speed up with satisfactory accuracy when compared against Celsius 3D, a FEM-based commercial solver.
\end{itemize}





\section{Background: Thermal simulation in 3D IC}
Here we provide a brief overview of thermal simulation in the context of 3D IC design. Thermal simulation aims to predict the temperature field of a given object (chip) $S$ by solving the heat conduction PDE globally. The 3D governing PDE is written as
\begin{equation}
\frac{\partial}{\partial y_1}\left(k \frac{\partial T}{\partial y_1}\right)+\frac{\partial}{\partial y_2}\left(k \frac{\partial T}{\partial y_2}\right)+\frac{\partial}{\partial y_3}\left(k \frac{\partial T}{\partial y_3}\right)+q_V=\rho c_p \frac{\partial T}{\partial t},
\label{heat:general}
\end{equation}
where $T$ and $q_V$ represent the temperature and the rate of internally generated energy per unit volume at any spatial-temporal location $(y_1, y_2, y_3, t)\equiv (\boldsymbol{y}, t)$. Here $k, \rho, c_p$ are material-specific properties of $S$ denoting material conductivity, mass density, and heat capacity, respectively. 

We focus on the static temperature field for isotropic materials (i.e., $k_{y_1}$=$k_{y_2}$=$k_{y_3}$=$k$), and simplify \eqref{heat:general} by setting $\frac{dT}{dt}=0$:
\begin{equation}
    k\cdot \nabla^2 T + q_V = 0,
    \label{heat}
\end{equation}
in which $\nabla^2$ stands for the laplacian operator. We then solve \eqref{heat}, with appropriately defined boundary conditions in the context of 3D IC design, for various chip designs
to find the optimal design by thresholding the temperature field.

\section{Modular Chip Configurations for Thermal Analysis}
\label{sec:modular}
\begin{figure}
    \centering
    \includegraphics[width=\columnwidth]{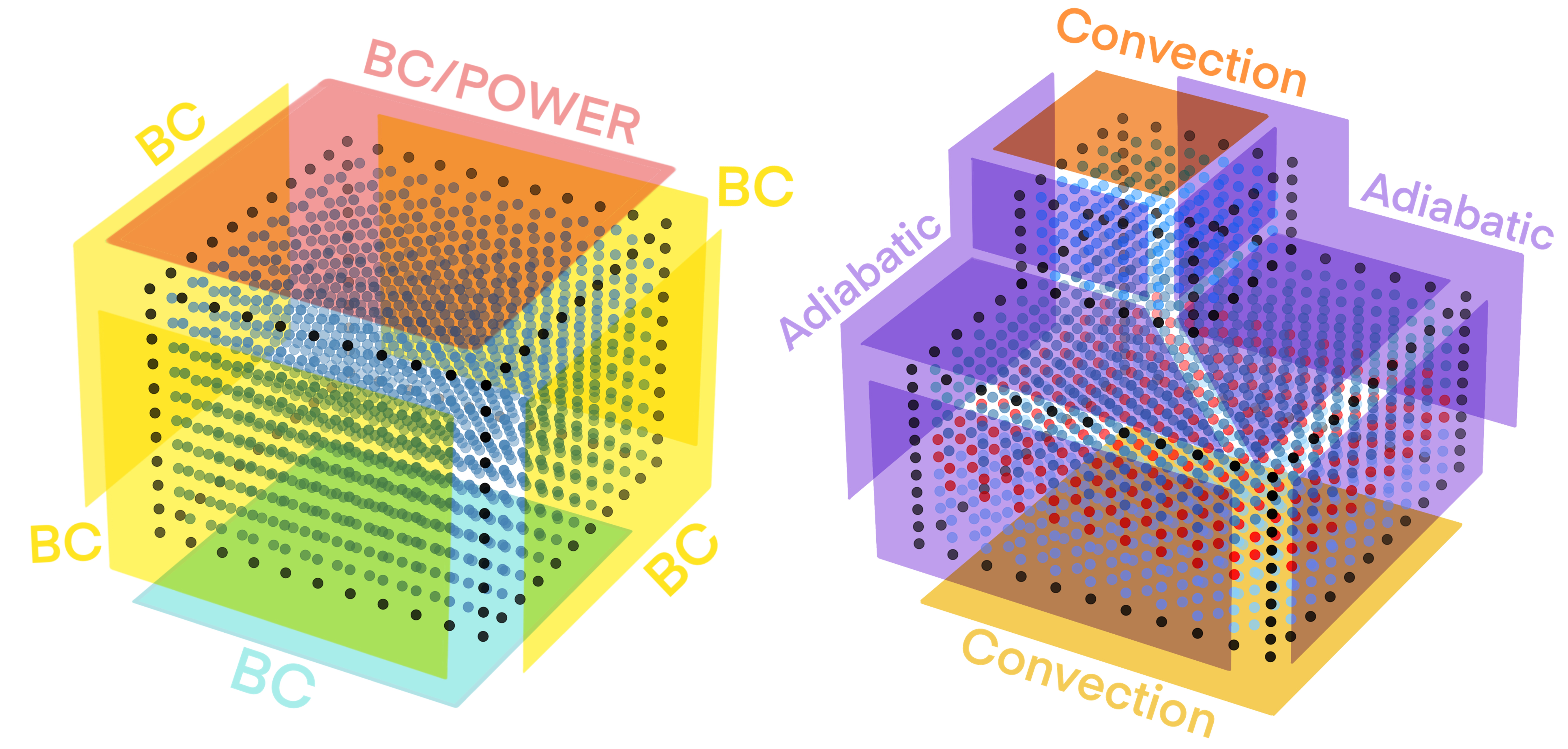}
    \caption{Schematic figures of chip designs in thermal simulation. The left one shows a general single cuboid chip model, of which the right one is a concrete implementation.}
    \label{fig:setting}
    \vspace{-10pt}
\end{figure}

Without loss of generality, we model the geometry of a chip as single or multiple stacked rectangular cuboid(s) as shown in Fig.~\ref{fig:setting}. For each cuboid, its temperature field depends on some key design configurations which include, but are not limited to, material/geometric parameters.

The first family of design configurations is the boundary condition (BC) for each individual surface that is exposed to the environment. We consider the following types of BCs:
\begin{itemize}[leftmargin=*]
    \item {\bf Dirichlet}: the temperature field on a surface is fixed as $q_d$:
    \begin{equation}
    T = q_d.
    \label{bc:dirichlet}
    \end{equation}
    \item {\bf Neumann}: the temperature flux on a surface is fixed
    \begin{equation}
        -k\frac{\partial T}{\partial y_i} = q_n,
        \label{bc:neumann}
    \end{equation}
    where $q_n$ represents the local heat flux density at the surface. 
    \item {\bf Adiabatic}: a special case of Neumann BC when $q_n$ is $0$ everywhere. This indicates a perfectly insulated surface.
    \item {\bf Convection}: also known as Newton BC.  This BC corresponds to a balance between heat conduction and convection in the same direction at the surface:
    \begin{equation}
        -k\cdot \frac{\partial T}{\partial y_i} = h(T-T_{\rm amb}).
        \label{bc:newton}
    \end{equation}
Here $h$ and $T_{\rm amb}$ stand for the heat transfer coefficient at the surface and the ambient temperature.
\end{itemize}

The second family of key design configurations are the locations and intensity of external/internal heat sources. This work considers the following two types of heat sources:
\begin{itemize}[leftmargin=*]
    \item {\bf Surface/2D power}: defined by the Neumann BC \eqref{bc:neumann} when $q_n$ is positive somewhere. Such $q_n$ is referred to as a surface/2D power map.
    \item {\bf Volumetric/3D power}: defined by the heat equation \eqref{heat} when $q_V$ is positive somewhere. Such $q_V$ is referred to as a volumetric/3D power map.
\end{itemize}

We now present the thermal chip designs by several independent modular configurations as shown in Fig.~\ref{fig:setting}. The left figure shows a general single cuboid chip with different BCs defined on each surface. The BC for the top surface also defines a 2D power map. The uniform blue color for the dots inside the cuboid indicates homogeneously distributed conductivity without any internal heat source. As a comparison, the right figure indicates a concrete implementation. In this model, we have volumetric power shown as the red dots in the middle layer of the bottom cuboid with adiabatic BCs on all side surfaces and convection BCs on the top and bottom surfaces. The different colors applied to the convection surfaces and the internal blue dots indicate different heat transfer coefficients and inhomogeneously distributed conductivity.

The above design configurations can change the PDE structure and temperature field of a 3D IC significantly. Many of them are described as functions instead of parameters, and they cannot be handled by traditional machine learning techniques.

\begin{figure*}
\vspace{-10pt}
    \centering
    \includegraphics[width=\textwidth]{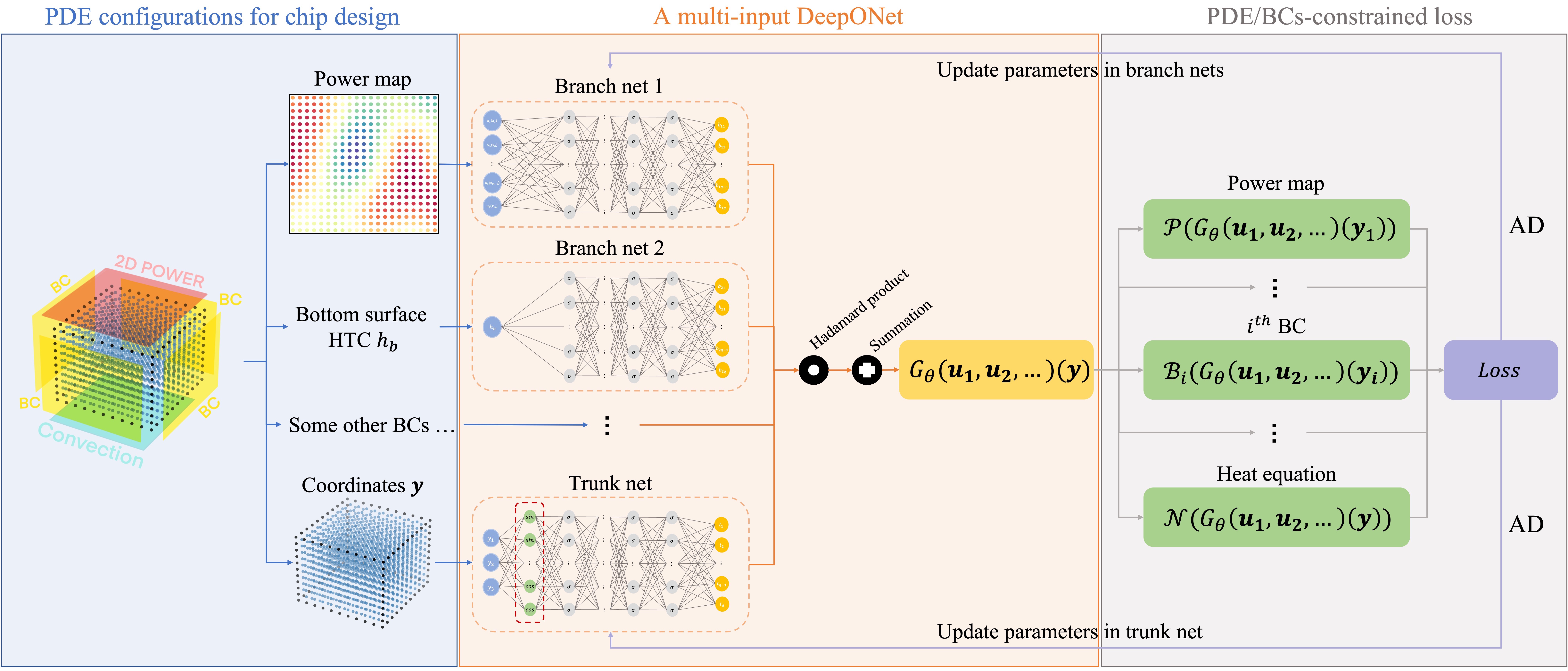}
    \caption{The proposed DeepOHeat framework.}
    \label{fig:DeepOHeat}
    \vspace{-10pt}
\end{figure*}

\section{The DeepOHeat framework}

 Now we present DeepOHeat: a self-supervised operator learning-based neural thermal solver enabling ultra-fast thermal prediction. DeepOHeat takes functions that characterize key design configurations (e.g., power maps, boundary conditions, domain of interest) rather than just material or geometric parameters as inputs to predict temperature fields in real time. The key ideas of DeepOHeat are shown in Fig.~\ref{fig:DeepOHeat}.

\subsection{Learning the Solution Dependence on Multiple PDE Configurations via a Multi-input DeepONet}
For succinct notations, we denote the heat equation of interest \eqref{heat} in the following general format
\begin{equation}
    \mathcal{N}(\boldsymbol{s} (\boldsymbol{u}_1, \boldsymbol{u}_2, \ldots, \boldsymbol{u}_k)(\boldsymbol{y})) = 0.
    \label{pde}
\end{equation}
Here $\mathcal{N}$ is a symbolic representation of the simplified heat equation \eqref{heat}. We denote the temperature field, i.e., the solution function of this PDE, as $\boldsymbol{s}$. A concrete temperature field is determined by a certain chip design specified by various configurations such as a power map and BCs. We present in general $k$ design configurations of interest (i.e., PDE configurations), both parametric and non-parametric, as $\boldsymbol{u}_1, \boldsymbol{u}_2, \ldots, \boldsymbol{u}_k$. Given specific PDE configurations $\{\boldsymbol{u}_i\}_{i=1}^k$, the temperature field on the domain of interest is then evaluated on the corresponding spatial coordinates $\boldsymbol{y}$, yielding the final formal representation as $\boldsymbol{s} (\boldsymbol{u}_1, \boldsymbol{u}_2, \ldots, \boldsymbol{u}_k)(\boldsymbol{y})$.

To avoid any potential confusion, we emphasize that each $\boldsymbol{u}_i$, $i=1, 2, \ldots, k$, no matter which representation form it uses, is represented as a \textbf{function} instead of a parameter in DeepOHeat. Therefore, DeepOHeat is designed to learn a functional map $G_{\boldsymbol{\theta}}$ ($\boldsymbol{\theta}$ denote all the neural network parameters in DeepOHeat, i.e., weights and bias) that maps the function space spanned by the PDE configurations $\{\boldsymbol{u}_i\}_{i=1}^k$, denoted by $\mathcal{U}: \mathcal{U}_1 \times \mathcal{U}_2 \times \cdots \times \mathcal{U}_k$, to the corresponding function space $\mathcal{S}$ spanned by its temperature field $\boldsymbol{s} (\boldsymbol{u}_1, \boldsymbol{u}_2, \ldots, \boldsymbol{u}_k)(\boldsymbol{y})$, i.e.,
\begin{equation}
    G_{\boldsymbol{\theta}}: \mathcal{U} \rightarrow \mathcal{S}.
\end{equation}
Such a map means that, a well-trained DeepOHeat is capable of accurately predicting the temperature field given any unseen design drawn from the same PDE configurations space $\mathcal{U}$. To learn this functional map, we leverage recent works in operator learning, DeepONets~\cite{deeponet} and multi-input DeepONets~\cite{mionet}.

{\bf Encoding Design Configurations as Input Functions of DeepOHeat.} We consider the general case that $k$ PDE configurations are considered. Correspondingly, we will have $k$ different input functions. For the $i^{th}$ configuration, we consider a random sample $\boldsymbol{u}_i^{(j)}$ drawn from its function space $\mathcal{U}_i$. This function (e.g., a 2D power map) is identified by its values on fixed locations $(\boldsymbol{x}_1, \boldsymbol{x}_2, \ldots, \boldsymbol{x}_m)$ (e.g., some grid points of a surface), and is then fed as an $m$-dimensional vector into the $i^{th}$ sub-network block, namely the $i^{th}$ ``branch net"~\cite{deeponet}. Repeating this process for all $k$ design configurations, we then have $k$ different input functions and the corresponding branch nets. All these configurations are from a certain design thus share the same domain of interest.  we then input all the coordinates sampled from this simulation domain into another sub-network, namely the "trunk net". To effectively learn the high-frequency information of the temperature field, we also apply a Fourier features mapping \cite{ffn} to the first layer of the trunk net, which is shown inside the dashed red box in the trunk net part of Fig.~\ref{fig:DeepOHeat}. 

{\bf Example.} We consider the example shown in the left part of Fig.~\ref{fig:DeepOHeat}. We see that for this single-cuboid chip, we define a 2D power map on the top surface. The power map that can have an arbitrary layout of heat sources, is with no doubt a non-parametric function. We identify this 2D power map by its values on equispaced grid points, which naturally form a two-dimensional matrix as shown in Fig.~\ref{fig:DeepOHeat}. We then flatten this matrix to a vector and feed it into the first branch net. If we consider a 3D power map, everything will be exactly the same except it will be identified by its values on three-dimensional equispaced grid points. Meanwhile, we define a convection BC on the bottom surface of the chip with a uniform HTC distribution of value $h_b$. In this case, the HTC on the bottom surface can be seen as a constant function therefore only one grid point is needed to identify this configuration. We then input $h_b$ into the second branch net. Note that $h_b$ should still be regarded as a function that has a parametric format instead of a parameter. If the surface has an inhomogeneous HTC distribution, one can simply encode it similarly as we encode a 2D power map. For the side surfaces of the chip, other BCs are defined accordingly and encoded as other DeepOHeat inputs or just fixed invariant configurations.

With $k$ defined PDE configuration inputs and the domain coordinates, we have in total $k$ branch nets and one trunk net, each of which outputs a $q$-dimensional feature vector. We then follow the ideas in \cite{mionet} to combine all these output features via Hadamard (element-wise) product and then sum up the resulting vector to a scalar output that represents the predicted temperature field, denoted as $T=G_{\boldsymbol{\theta}}(\boldsymbol{u}_1, \boldsymbol{u}_2, \ldots, \boldsymbol{u}_k)(\boldsymbol{y})$.

\subsection{Training DeepOHeat via Physics-Informed Loss}
Now we explain how to train the DeepOHeat network. According to \cite{deeponet}, a DeepONet is generally trained via a data-driven approach, in which data triplets $(\boldsymbol{y}, \{\boldsymbol{u}_i\}_{i=1}^k, \boldsymbol{s})$ need to be collected via massive runs of numerical simulation. 
For relatively complicated chip designs, a single FEM simulation might cost hours or even days to complete. Therefore, large-scale data collection is practically prohibitive in this context. Instead, we follow the idea from a recent approach~\cite{pideeponet}, which leveraged the ideas from physics-informed neural networks (PINNs) \cite{pinns} to train a single-input DeepONet for solving parametric PDEs. We extend their work to handle multi-input scenarios as shown on the right of Fig.~\ref{fig:DeepOHeat}.

Again we consider the aforementioned general case where $k$ chip design configurations are considered. For the $i^{th}$ configuration $\boldsymbol{u}_i$, we first index all the coordinates that are located in its designated regions, such as a boundary surface, denoted as $\boldsymbol{y}_i$. Then on $\boldsymbol{y}_i$,  we impose a physics constraint $\mathcal{L}_i$. If $\boldsymbol{u}_i$ represents a power map, we denote $\mathcal{L}_i$ as
\begin{equation}
    \mathcal{L}_i = \left\|\mathcal{P}\left(G_{\boldsymbol{\theta}}(\boldsymbol{u}_1, \boldsymbol{u}_2, \ldots, \boldsymbol{u}_k)(\boldsymbol{y}_i)\right)\right\|.
\end{equation}
For a 2D power map, $\mathcal{P}$ is a symbolic representation of the Neumann BC \eqref{bc:neumann}. For a 3D power map, $\mathcal{P}$ will represent the heat equation \eqref{heat} with non-zero $q_V$. If $\boldsymbol{u}_i$ represents a general BC, such as convection or Dirichlet BC, we denote $\mathcal{L}_i$ as
\begin{equation}
    \mathcal{L}_i = \left\|\mathcal{B}_i\left(G_{\boldsymbol{\theta}}(\boldsymbol{u}_1, \boldsymbol{u}_2, \ldots, \boldsymbol{u}_k)(\boldsymbol{y}_i)\right)\right\|,
\end{equation}
where $\mathcal{B}_i$ denotes the formulation of the corresponding BC. For the entire domain of interest, we impose the PDE constraint, except for the region where a 3D power map is imposed, as
\begin{equation}
    \mathcal{L}_r = \left\|\mathcal{N}\left(G_{\boldsymbol{\theta}}(\boldsymbol{u}_1, \boldsymbol{u}_2, \ldots, \boldsymbol{u}_k)(\boldsymbol{y})\right)\right\|.
\end{equation}
We then obtain the total loss as 
\begin{equation}
    \mathcal{L}_{\text{total}} = \mathcal{L}_r + \sum_{i=1}^k \mathcal{L}_i.
\end{equation}
We train DeepOHeat by minimizing the total loss via gradient descent based on automatic differentiation algorithms \cite{ad}.

\begin{figure*}
\vspace{-20pt}
    \centering
    \includegraphics[width=\textwidth]{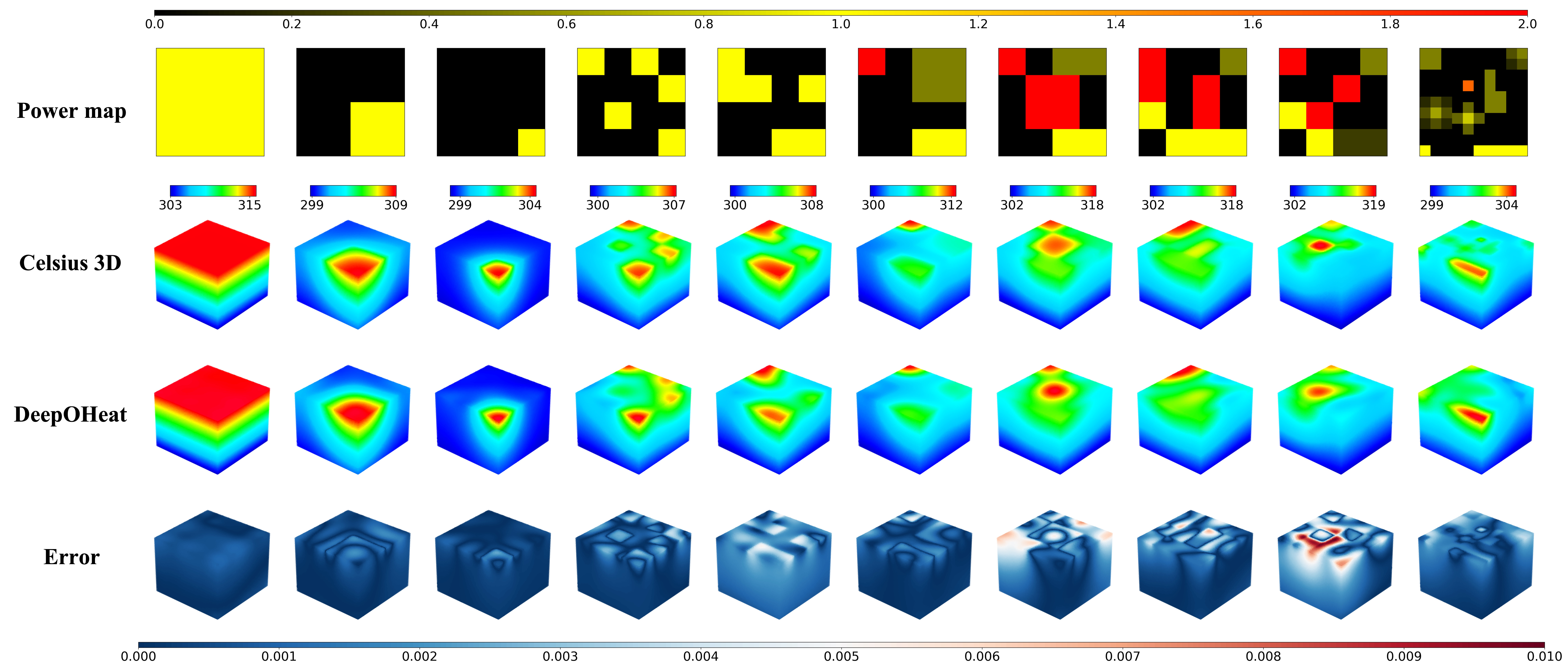}
    \caption{Predicted temperature fields for different 2D power maps defined on the top surface.}
    \vskip -0.2in
    \label{fig:2d power map}
\end{figure*}

\section{Experiments}
In this section, we present two implementations of the proposed DeepOHeat and compare our results with Celsius 3D, a state-of-the-art numerical solver for 3D chip thermal analysis from Cadence Design Systems. Our results demonstrate that, for any unseen designs, a well-trained DeepOHeat is capable of producing satisfactory results with at least 1000$\times$ speedup.

\subsection{2D Power Map Configuration on The Top Surface}
As the power map controls the heat generation in a certain chip design, the prediction performance of DeepOHeat on unseen new power maps are of major interest. For illustration, here we focus solely on optimizing a 2D power map by training a single-input DeepOHeat.

\begin{figure}[t]
    \centering
    \includegraphics[width=\columnwidth]{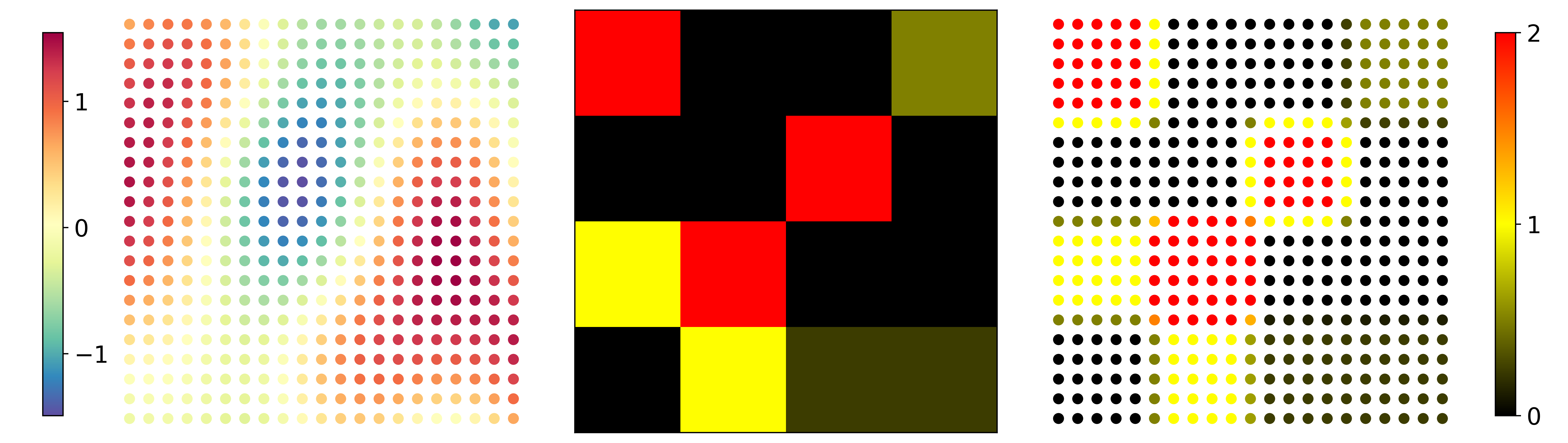}
    \caption{Left: a power map for training; Middle and right: test power maps for Celsius 3D and for DeepOHeat, respectively.}
    \label{fig:power maps}
\end{figure}

\subsubsection{Problem setup}
We consider a $21\times 21 \times 11$ mesh grid-based single-cuboid geometry which represents a $1\text{mm} \times 1\text{mm} \times 0.5\text{mm}$ chip in practice. This geometry is similar to the one shown in the left of Fig.~\ref{fig:setting} and has in total of 4851 grid points. We define a 2D power map on the top surface, in which a one-unit power corresponds to a $0.00625(mW)$ power in real-world settings. We define Adiabatic BC on all side surfaces and convection BC on the bottom surface with $\text{HTC}=500 W/(m^2 K)$ and $T_{\rm amb}=298.15 (K)$. A homogeneous thermal conductivity $k = 0.1 W/(mK)$ is assigned to the entire domain and no volumetric power is applied.

\subsubsection{Generating training power maps}
We sample all the training power maps from a two-dimensional standard Gaussian random field (GRF) with the length scale parameter equal to 0.3. The length scale controls the smoothness of the sampled functions. We choose 0.3 in this example to generate relatively smooth power maps as shown on the left of Fig.~\ref{fig:power maps}. One can also tune this parameter to generate training power maps similar to those in specific optimization tasks. Corresponding to our $21 \times 21$ mesh grids on the top surface, we identify each power map by its values on these coordinates formatted as a matrix of the same size. We then flatten these matrices to vectors of length 441 as the input of the branch net.

\subsubsection{DeepOHeat settings}
In this example, we use a 9-layer branch net with 256 neurons per layer combined with a 6-layer trunk net with 128 neurons per layer. The first layer of the trunk net is a Fourier features mapping \cite{ffn} where its coefficients are sampled from a normal distribution with zero mean and $2\pi$ standard deviation. The input dimensions of the branch net and the trunk net are 441 and 3, which correspond to the dimensions of the encoded power map and the 3D spatial coordinates, respectively. The output dimensions of the two sub-networks are both 128. We set all the activation functions as the "Swish" function proposed by Ramachandra et al.\cite{swish}. We find in experiments that Swish yields relatively better results compared to other popular activation functions used in PINNs, such as Sine and Tanh.

\subsubsection{Training settings}
We train this DeepOHeat by 10000 iterations to guarantee convergence, which takes 10 hours on a single Tesla V100 GPU. In each iteration, 50 input functions are sampled from the given GRF and fed into the branch net. For each function, the 4851 mesh grid points of the entire simulation domain are fed into the trunk net. We therefore have a $242550 \times 441$ input for the branch net and a $242550\times 3$ input for the trunk net. We choose the initial learning rate as 1e-3 and decay the learning rate by $0.9\times$ every 500 iterations.

\subsubsection{Test settings}
We aim to compare the predicted temperature fields with Celsius 3D element-wisely on unseen new power maps. There exists a minor discrepancy between the power maps of Celsius 3D and DeepOHeat. As shown in the middle of Fig.~\ref{fig:power maps}, the power maps in Celsius 3D are tile-based, different from the grid-based ones in DeepOHeat. To accommodate these realistic power maps used in Celsius 3D, we interpolate the $20\times 20$ tile-based power maps to $21\times 21$ grid-based power maps, as shown in the middle and right of Fig.~\ref{fig:power maps}. Such a transformation not only enables DeepOHeat to accept almost the same realistic power maps as in Celsius 3D but also smooths out these discretely defined power maps. As the ones we used for training are all continuous functions, using the smoothed rather than discrete power maps for testing, in a heuristics sense, would have lower generalization errors.

\subsubsection{Results}
\begin{table}[t]
\caption{Mean and peak errors for all power maps}
\begin{center}
\scalebox{0.8}{
\begin{tabular}{c|c|c|c|c|c|c|c|c|c|c}
\toprule
& $p_1$ & $p_2$ & $p_3$ & $p_4$ & $p_5$ & $p_6$ & $p_7$ & $p_8$ & $p_9$ & $p_{10}$ \\ \midrule
MAPE (\%) & 0.03 & 0.03 & 0.02 & 0.05 & 0.14 & 0.04 & 0.13 & 0.07 & 0.16 & 0.08 \\ \midrule
PAPE (\%) & 0.10 & 0.20 & 0.24 & 0.38 & 0.52 & 0.49 & 0.71 & 0.66 & 1.00 & 0.40  \\ \bottomrule
\end{tabular}}
\vspace{-20pt}
\label{tab:error 2d power map}
\end{center}
\end{table}

All the results of this example are shown in Fig.~\ref{fig:2d power map}. From the left to the right, we gradually increase the complexity of the unseen test power maps. We simply refer them to as $p_1, p_2, \ldots, p_{10}$ and report their mean absolute percentage errors (MAPEs) and peak absolute percentage errors (PAPEs) in Table.~\ref{tab:error 2d power map}. We see that DeepOHeat is capable of predicting the temperature fields for all these unseen test power maps with satisfactory accuracy. Moreover, we want to highlight the strong generalization power of DeepOHeat. All these power maps, though most of which are composed of heat blocks, are quite different from the training power maps. Specifically, the last power map $p_{10}$ can be seen as a very wiggly function in this context. We see that $p_{10}$ has multiple small-sized heat sources and one of them is also given a relatively large power. For such a complicated power map, DeepOHeat still yields satisfactory predictions at the most part of the domain, except for mildly overestimated temperatures at the regions between those small-sized heat sources.

\subsubsection{Speedup} In this example, Celsius 3D costs approximately 5min for a single simulation on an Intel Xeon Gold 6148 CPU. The post-training prediction time for DeepOHeat is 0.1s on the same CPU and 0.001s on a Tesla V100 GPU, which correspond to a 3000$\times$ and 300000$\times$ speedup, respectively. For a larger-scale or more complicated design, the computational cost for FEM-based solvers will rapidly increase while remaining unchanged for DeepOHeat. We expect more significant speed-up in realistic thermal optimization tasks.

\subsection{HTC Configurations on Both Top and Bottom Surfaces}

\begin{figure}
\vspace{-10pt}
    \centering
    \includegraphics[width=\columnwidth]{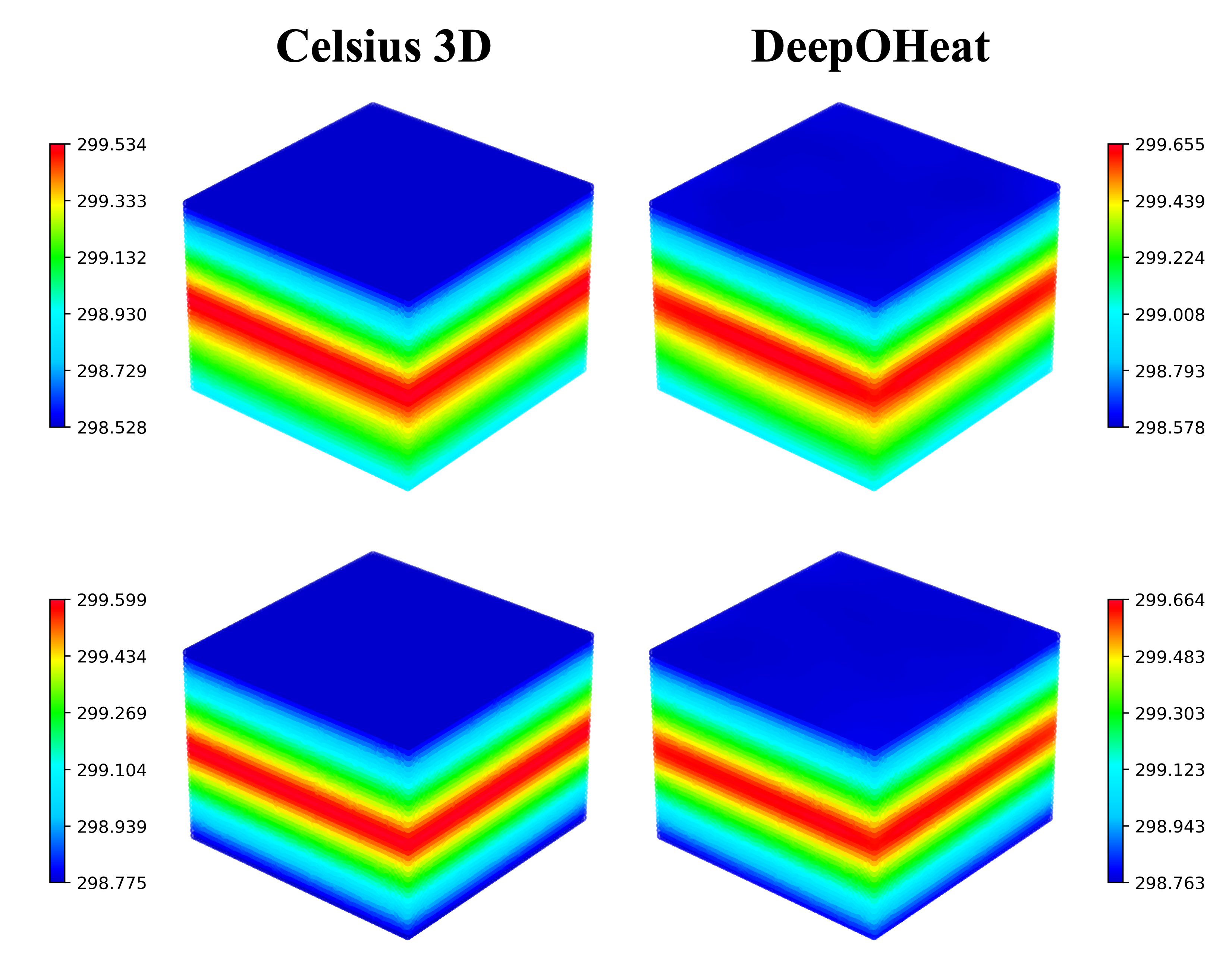}
    \caption{Temperature fields under different HTC configurations.}
    \label{fig:results htc}
    \vspace{-15pt}
\end{figure}

DeepOHeat can predict the thermal behaviors influenced by multiple design configurations. To demonstrate this, we build a dual-input DeepOHeat to predict the temperature field of a 3D IC influenced by the HTCs on two surfaces simultaneously. In this example, we avoid introducing detailed settings instead focus on those that are different from the previous example. 

We consider a similar cuboid chip geometry with the size of $1\text{mm} \times 1\text{mm} \times 0.55\text{mm}$ but all 7000 points are randomly sampled inside (on) the entire domain. We don't use mesh because we don't have mesh-based encoding for this example. We define convection BCs for both top and bottom surfaces and assume HTCs are constantly distributed. We define a single-layer uniform volumetric power with a thickness of $0.05\text{mm}$ and the value of $0.000625 (W)$. The settings for side surfaces and thermal conductivity are the same as before.

In each iteration, we sample 20 i.i.d samples uniformly from a squared area $[333.33, 1000] \times [333.33, 1000] \ (W/m^2K)$, corresponding to 20 different HTCs for both two surfaces. For each sampled HTC tuple, we randomly draw a new set of coordinates from the simulation domain. Combining these, we have two $140000 \times 1$ inputs for the two branch nets and a $140000 \times 3$ input for the trunk net. 

In this example, we use relatively simpler networks for the two branch nets, each of which contains 5 fully-connected layers with only 20 neurons per layer. The trunk net still has 6 layers with 128 neurons per layer and a Fourier features mapping defined in the first layer with a $\pi$ standard deviation this time. The output dimensions for all sub-networks are 50.

After training DeepOHeat for 5000 iterations (about 2 hours), we evaluate its performance on some unseen values sampled from the same 2D region. For example, we pick two sets of HTCs, $(1000, 333.33)$ and $(500, 500)$, as the test cases and show the corresponding results in each row of Fig.~\ref{fig:results htc}. Although different HTCs make only slight differences, DeepOHeat still yields accurate predictions in both cases. As shown by the color bars in Fig.~\ref{fig:results htc}, the differences in the predicted maximal and minimal temperatures between Celsius 3D and DeepOHeat are within $0.1 (K)$. In the first case where $\text{HTC}=1000$ on the top surface and $\text{HTC}=333.33$ on the bottom surface (first row in Fig.~\ref{fig:results htc}), the MAPE and PAPE of DeepOHeat are $0.032\%$ and $0.043\%$. In the second case where $\text{HTC}=500$ on both two surfaces (second row in Fig.~\ref{fig:results htc}), the MAPE and PAPE of DeepOHeat are $0.011\%$ and $0.025\%$. 

Celsius 3D costs around 2min for a single simulation on the aforementioned CPU. The runtime for DeepOHeat remains unchanged. Therefore the speed up in this example is 1200$\times$ and 120000$\times$ on CPU and GPU, respectively.

\section{Conclusion}
In this work, \textit{for the first time}, we have introduced a physics-aware operator learning framework, named DeepOHeat, to perform ultra-fast 3D chip thermal prediction under multiple chip design configurations. We have proposed a modular chip thermal model to encode various chip geometries, power maps, and boundary conditions. We have applied a physics-informed multi-input DeepONet to seamlessly solve a family of heat equations that take multiple BCs and the power map as input configurations with no data supervision required. The experiments on two specific tasks show that a well-trained DeepOHeat can predict the temperature fields on unseen new chip designs with high accuracy while no noticeable simulation time is required. In the future, we will further investigate how DeepOHeat performs in more complicated geometries and in optimizing 3D power maps.

\small
\bibliographystyle{IEEEtran}
\bibliography{ref}
\end{document}